\documentclass[11pt]{article}
\usepackage[margin=1in]{geometry}
\usepackage{amssymb,amsmath,amsthm}
\usepackage{multirow}
\usepackage{graphicx}
\usepackage{hyperref}

\usepackage[numbers]{natbib}

\usepackage{authblk}

\begin{document}

\title{RobustDebias: Debiasing Language Models using Distributionally Robust Optimization}


\author[1]{Deep Gandhi}
\author[1]{Katyani Singh}
\author[1,2]{Nidhi Hegde}

\affil[1]{University of Alberta}
\affil[2]{Alberta Machine Intelligence Institute}







\maketitle
\begin{abstract}

 Pretrained language models have been shown to exhibit biases and social stereotypes. Prior work on debiasing these models has largely focused on modifying embedding spaces during pretraining, which is not scalable for large models. Fine-tuning pretrained models on task-specific datasets can both degrade model performance and amplify biases present in the fine-tuning data. We address bias amplification during fine-tuning rather than costly pretraining, focusing on BERT models due to their widespread use in language understanding tasks. While Empirical Risk Minimization effectively optimizes downstream performance, it often amplifies social biases during fine-tuning. To counter this, we propose \textit{RobustDebias}, a novel mechanism which adapts Distributionally Robust Optimization (DRO) to debias language models during fine-tuning. Our approach debiases models across multiple demographics during MLM fine-tuning and generalizes to any dataset or task. Extensive experiments on various language models show significant bias mitigation with minimal performance impact.

\end{abstract}



\section{Introduction}



Pretrained Language Models (PLMs), often commonly referred to as Large Language models (LLMs), are used for a variety of tasks such as text generation, summarization, translation, classification, etc. ~\cite{devlin2019bert,brownfewshot2020}. These models are trained on massive corpora collected from diverse sources~\cite{gao2020pile}, with their success largely attributed to transformer architectures~\cite{vaswani-attention-2017} that learn hierarchical representations and capture long-range contextual dependencies. Models like BERT (Bidirectional Encoder Representations from Transformers)~\cite{devlin2019bert} use bidirectional encoders to analyze words in full context by considering both preceding and succeeding words, while GPT-style models~\cite{brown2020language} use autoregressive decoders that generate predictions for the next word based on preceding context.

Research has demonstrated that these models inherit and amplify biases present in their training data~\cite{meade2022empirical}, reflecting societal, cultural, and linguistic prejudices embedded in text corpora. In the context of language models, bias manifests as unfair representation of certain groups, perspectives, or characteristics, often reinforcing historical stereotypes, privileging certain demographics, or marginalizing underrepresented groups. These biases are often inherited from the data used for training these language models~\cite{arteaga-bias-2019}. For instance, when presented with the sentence "The doctor presented [BLANK] research at the conference," language models disproportionately predict "his" over "her," reflecting gender-based stereotypes about medical professions~\cite{bolukbasi2016man}. Such biases propagate through pre-training and cause harm to marginalized groups when these models are deployed in real-world applications. As LLMs become increasingly sophisticated and continue to shape natural language processing (NLP), the ethical implications of these biases gain prominence, necessitating critical examination of their impact on downstream applications and the development of mitigation strategies that balance their remarkable capabilities with ethical considerations surrounding fairness.

Debiasing refers to the process of mitigating biases present within a system, dataset, or model. Prior works~\cite{bolukbasi2016man,ravfogel2020null,zmigrod2019counterfactual} focus on eliminating stereotypical associations in language models but their scope is typically limited to addressing one bias demographic at a time. Early approaches like Hard Debias~\cite{bolukbasi2016man}, proposed for word2vec embeddings~\cite{mikolov2013efficient}, focused on identifying bias subspaces and modifying word embeddings to remove their effects. Similar techniques such as GN-GLoVE~\cite{zhao2018learning}, INLP~\cite{ravfogel2020null}, and SENT-DEBIAS~\cite{liang2020towards} attempted to identify latent subspaces of biases in the embedding space and remove their effects. Some methods directly alter word embeddings (Hard Debias, GN-GLoVE) or contextual embeddings (SENT-DEBIAS), while others like INLP~\cite{ravfogel2020null} use series of classifiers to predict bias-free subspaces and sequentially project embeddings to remove bias attributes. However, works such as~\cite{bolukbasi2016man,ravfogel2020null,liang2020towards} that focused on identifying bias subspaces within representations may not ensure that downstream tasks or further fine-tuning reflect the debiased language. Additionally, using several different models each debiased for only one bias demographic is computationally inefficient in real-time scenarios. Other methods like Counterfactual Data Augmentation~\cite{zmigrod2019counterfactual} modify training data by augmenting it with counter-stereotypical examples, such as pairing "he is a doctor" with "she is a doctor" to mitigate gender bias caused by data imbalance.

More recent approaches have introduced additional techniques but maintain similar limitations. AutoDebias~\cite{guo2022auto} proposes a prompt-based approach for searching biases associated with target words (words representing biased subgroups, e.g., "man/woman", "husband/wife") and attribute words (words representing attributes for which subgroups are biased, e.g., "doctor/nurse" for professions). Given a list of concept words, prompts are synthetically created by using a concept word and selecting the most disagreeable word to fill the remaining space. For example, for a template "A [placeholder] has a job as [MASK]," concept words "he" and "she" are placed as placeholders, then the probability of attribute words is examined to select those with least probability, such as "housekeeper" and "receptionist" for "he" and "engineer" and "CEO" for "she". An objective function then minimizes the distance between probabilities of these synthetic prompts. 

CausalDebias~\cite{zhou-etal-2023-causal} identifies causal factors that amplify bias by finding sentences exhibiting gender bias based on target word lists, creating counterexamples by replacing target words ("he" with "she"), and using causal invariant loss functions to eliminate representation links corresponding to gender biases. PCGU (Partitioned Contrastive Gradient Unlearning)~\cite{yu-etal-2023-unlearning} focuses on gradient direction changes in masked language modeling. For a masked sentence (e.g., "The surgeon could not operate on [MASK] patient"), the authors calculate and store backpropagated gradients for advantaged group predictions ([MASK] = "his") and disadvantaged group predictions ([MASK] = "her"). By calculating cosine similarity between these stored gradients, they determine which weight vectors change most significantly, freeze other vectors, and use stochastic gradient-based learning to reduce advantaged group probabilities by fine-tuning only affected weight vectors. 

However, both AutoDebias and CausalDebias rely on predefined knowledge about target words and specific bias types to be mitigated during fine-tuning, which is not scalable as the number of bias types increases. PCGU focuses exclusively on gender bias scenarios. All these approaches depend on word lists or predefined knowledge about bias demographics. Manually defining these metrics for every possible bias type is very difficult to achieve. \textbf{Thus, there is a need for a robust approach that can mitigate biases for multiple demographics simultaneously without such dependencies.}

 Our work presents a method to mitigate commonly observed biases in language models during the fine-tuning phase of large language model training. In our approach, we replace the standard objective function with an alternative optimization approach. This change in the fine-tuning objective aims to address biases present in the model's pre-trained representations and subsequently in its generated outputs for any downstream task. The goal is to promote equal performance across multiple subgroups of people belonging to various bias demographics while preserving the actual language modeling performance of the model.

The language task we focus on is Masked Language Modeling (MLM), a technique used in language models, particularly in pre-training models such as BERT~\cite{devlin2019bert}. In MLM, certain tokens in the input sequence are randomly masked or replaced with a special token (typically [MASK]), and the model is trained to predict the original tokens based on context provided by surrounding tokens. This task encourages the model to learn contextualized representations of words, enabling language models to better capture nuanced meanings and to generate coherent and contextually relevant text. Continuing our example of the sentence "The doctor presented [BLANK] research at the conference," a perfectly debiased model in the absence of additional context should prefer the pronouns "his" and "her" equally during the inference phase.  We apply our approach to the MLM task in the fine-tuning phase.

We propose \textit{RobustDebias}, a novel approach for fine-tuning LLMs using a Distributionally Robust Optimization (DRO) objective. We fine-tune a pre-trained model using a DRO objective rather than the conventional Empirical Risk Minimization (ERM) objective. We focus on two research questions:
\begin{enumerate}
    \item Are Distributionally Robust Optimization approaches better for debiasing language models than the traditional Empirical Risk Minimization objective? 
    \item Can we devise a debiasing approach which does not rely on external word lists and can debias multiple bias demographics at once effectively? 

\end{enumerate}

These research questions are motivated by two key observations: (1) DRO has shown promise in reducing spurious correlations in tasks such as natural language inference and object recognition
recognition ~\cite{sagawa2019distributionally}, but its application to bias mitigation in language tasks remains unexplored; and 
(2) existing debiasing methods either require manual curation of word lists for each bias type or can only address one demographic at a time, making them impractical for real-world deployment where multiple biases coexist. 

Our contributions are twofold. First, to the best of our knowledge, this is the first work that investigates the debiasing of an LLM when optimized using DRO. To this end, we adapted existing DRO frameworks previously employed for various Natural Language Processing (NLP) tasks to the debiasing context. Second, we introduce a novel DRO framework, \textit{RobustDebias}, which focuses on debiasing LLMs through fine-tuning and can be readily adapted to any dataset. Through our comprehensive experiments, we observe that \textit{RobustDebias} performs in a highly stable manner across different models. We compare \textit{RobustDebias} to several existing DRO approaches, the traditional ERM objective, as well as existing debiasing methods. Our results demonstrate that \textit{RobustDebias} achieves superior performance across all bias demographics. The ability of \textit{RobustDebias} to debias multiple bias demographics without its dependence on predefined word lists makes it a substantially more robust debiasing method for real-world deployment scenarios.

\section{Background}
\label{background}
In this section, we provide background on BERT language models, explain the principles of DRO and review existing DRO frameworks applied to NLP tasks.

\subsection{BERT language models}
\label{sec:llm-bg}

In recent years, the field of natural language processing has made significant advances to develop bigger and better autoregressive language models. These models are trained using the task of generating the next word in a sequence based on the previous words provided as an input. Among these models, BERT~\cite{devlin2019bert} demonstrated remarkable performance gains across various NLP tasks. The major advantage of the BERT model is the ease-of-use for different downstream tasks by using the pretrained representations and training them for adapting to the new task. This is usually referred to as fine-tuning. The BERT contextualized embeddings have been pretrained on several large datasets for the tasks of masked word prediction\footnote{This task is also referred as Masked Language Modeling} and Next Sentence Prediction (NSP). Given a tokenized input sequence $\mathbb{X} = \{x_1,x_2,x_3,\ldots,x_n\}$, BERT computes contextualized embeddings $\mathbb{H} = \{h_1,h_2,h_3,\ldots,h_n\}$. Since these embeddings capture rich contextual information, BERT can be used for a variety of downstream tasks by modifying task-specific layers in the model architecture and using a task-specific loss. 

Notable BERT variants include RoBERTa~\cite{liu2019roberta}, ALBERT~\cite{lan2019albert}, and DistilBERT~\cite{sanh2019distilbert}, which follow similar fine-tuning procedures but have their own unique characteristics and advantages. RoBERTa trains on dynamically sampled, larger corpora with longer sequences and uses dynamic masking patterns per training epoch. ALBERT reduces model parameters through parameter sharing across layers, enabling larger models on limited resources. DistilBERT uses knowledge distillation to mimic BERT with half the parameters.

In this work, we focus only on these encoder-only BERT models because: (1) they produce contextualized embeddings directly used in downstream tasks, making representation-level bias particularly consequential; and (2) as seen in~\cite{devlin2019bert}, the encoder based models can be used for a variety of downstream tasks such as sentiment analysis, feature extraction, and text generation. Such models are highly capable of capturing semantic meaning when being trained on diverse data. Fine-tuning these models for downstream tasks has also proven highly effective, as observed by~\cite{howard2018universal,cer2018universal}.

\subsection{Distributionally Robust Optimization}
\label{sec:dro-erm-detail}
Empirical Risk Minimization (ERM) is generally used as the training objective for large language models. The basic assumption in ERM is that since all training examples and test examples are independent and identically distributed (i.i.d.), the objective function focuses on minimizing the average loss on the training data as a whole. However, this may not always be helpful as seen by \cite{tu-etal-2020-empirical} where training using ERM develops some spurious correlations for majority classes in the training data. The objective function for updating model parameters ($\theta$) using ERM can be given as:
\begin{equation}
    \theta' = \arg\!\min_{\theta \in \Theta}\mathbb{E}_{(x,y) \sim \hat P}[\ell(f(x;\theta),y)],
\end{equation}

where $\ell$ is a given loss function, $f(x;\theta)$ is the prediction from a model with parameters $\theta$ given input $x$, $y$ represents the ground-truth, and $\hat P$ is the empirical distribution of the training data points.

Distributionally Robust Optimization (DRO) focuses on minimizing the worst-case expect loss in an uncertainty set of distributions $\mathcal{Q}$~\cite{ben2013robust}:

\begin{equation*}
    \theta' = \arg\!\min_{\theta \in \Theta}\sup_{Q \in \mathcal{Q}}\mathbb{E}_{(x,y) \sim Q}[\ell(f(x;\theta),y)],
\end{equation*}
The uncertainty set $\mathcal{Q}$ is typically chosen to include test distributions on which we want our model to perform well.  In the setting of subgroups within the data population, the uncertainty set is defined in terms of the groups, so that $\mathcal{Q}$ is some given mixture of group distributions~\cite{oren2019distributionally}. The training optimisation is then given by:  
\begin{equation}
    \theta' = \arg\!\min_{\theta \in \Theta}\max_{g \in \mathcal{G}}\mathbb{E}_{(x,y)\sim P_g}[\ell(f(x;\theta),y)],
\end{equation}
where $\mathcal{G}$ is the set of subgroups and $P_g$ is the empirical distribution of data points belonging to subgroup $g$.  
As we explain in Section~\ref{sec:proposed method}, for the debiasing application we define $P$ based on bias demographics present in training sentences. For example, a sentence containing gendered language belongs to the gender subpopulation, while one mentioning religious terms belongs to the religion subpopulation. 

\subsection{DROs in language models} 
Many notable works~\cite{oren2019distributionally,sagawa2019distributionally,broscheit-etal-2022-distributionally,xie2023doremi} have used DROs in language models for several tasks.  The use of a Conditional Value-at-Risk (CVaR) based DRO approach has been proposed~\cite{oren2019distributionally} to generate reviews on several topics. This method uses a topic model to obtain the distribution of sentences across different topics. The losses for every topic are collected at the end of every update and the final loss $\mathcal{L}_{\text{batch}}(\theta)$ is calculated as the average of losses across every topic, while only retaining losses that lie in the upper $\alpha$ percentile for that topic\footnote{For our experiments, we find the value of $\alpha=0.8$ to be most optimal value following a hyperparameter search.}. Mathematically, this can be expressed as:
\begin{align}
&\text{UpperPercentile}(t, \alpha) = \text{Percentile}(l_t, \alpha) \\
&\text{SelectedLosses}_t = \{ (x, y) \mid (x, y) \in \text{Batch} \land \text{Topic}(x) = t \nonumber \\
&\qquad\qquad\qquad\qquad\land l_{\text{task}}(x, y; \theta) \geq \text{UpperPercentile}(t, \alpha) \} \\
&\mathcal{L}_{\text{batch}}(\theta) = \sum_{t \in \mathcal{T}} \sum_{(x, y) \in \text{SelectedLosses}_t} l_{\text{task}}(x, y; \theta),
\end{align}

where $l_t$ represents the accumulated losses for topic $t$. $\text{UpperPercentile}(t, \alpha)$ computes the top $\alpha$ percentile of accumulated losses for topic $t$. $\text{SelectedLosses}_t$ represents the subset of losses selected from the topic $t$ based on the criterion described. For each topic $t$, only the losses which belong to the upper $\alpha$ percentile are selected. $\mathcal{L}_{\text{batch}}(\theta)$ computes the total loss using the selected losses.

\citet{sagawa2019distributionally} use the Group DRO algorithm for multi-genre natural language inference in addition to other vision-based tasks. For this approach, the authors focus on having pre-defined partitions of training data within the subgroups. This partition could be according to predefined labels, topics, etc. 
In this method, the batch losses are first averaged per-group and stored during training. The final loss $\mathcal{L}_{\text{batch}}(\theta)$ is calculated as the weighted average of these group losses.
The loss function  is formulated as follows~\cite{sagawa2019distributionally}:
\begin{align}
\mathcal{L}_{\text{group}}(g; \theta) &= \mathbb{E}_{(x, y) \sim \mathcal{D}_g} \left[ l_{\text{task}}(x, y; \theta) \right] \\
\mathcal{L}_{\text{batch}}(\theta) &= \sum_{g \in \mathcal{G}} \frac{b_g}{b} \mathcal{L}_{\text{group}}(g; \theta),
\end{align}

where $\mathcal{G}$ denotes the set of groups. $g\in\mathcal{D}_g$ denotes the individual groups, $\mathcal{D}_g$ denotes the distribution of examples belonging to group g, 
$b_g$ denotes the frequency of group $g$ and $b$ is the batch size.

Unlike Group DRO and Topic CVaR, no subpopulation information is necessary for TopK DRO ~\cite{levy2020large,kawaguchi2020ordered}. The authors simply define the final loss $\mathcal{L}_{\text{batch}}$ in TopK DRO as the average of the top-k largest losses in a batch for every update. This can be expressed as follows:
\begin{align}
\mathcal{L}_{\text{batch}}(\theta) &= \frac{1}{k} \sum_{i=1}^{k} \text{SortDescending}(\{l_{\text{task}}(x_i, y_i; \theta) \mid (x_i, y_i) \in \text{Batch}\}),
\end{align}
where $k$ is the number of losses to consider. $(x_i, y_i)$ represents the i-th example in the batch. $l_{\text{task}}(x_i, y_i; \theta)$ represents the loss for that i-th example and $\text{SortDescending}(...)$ arranges the accumulated losses in a descending order.

~\citet{broscheit-etal-2022-distributionally} use these existing DRO approaches and also introduce two new approaches for spoken language understanding tasks such as intent classification and slot filling. One of these new approaches focuses on combining the Group DRO approach~\cite{sagawa2019distributionally} with a Top-K approach~\cite{levy2020large,kawaguchi2020ordered} to only consider the largest losses within subgroups. In the proposed TopK-Group DRO method, the predefined subpopulation information is required for fine-tuning. However, unlike the Group DRO algorithm, only the top-k largest losses are selected to be included in the average in each group. The final loss $\mathcal{L}_{\text{batch}}$ is then taken as the loss of the group with the largest top-k loss. It can be expressed as follows:
\begin{align}
\mathcal{L}_{\text{group}}(g; \theta) &= \frac{1}{k} \sum_{i=1}^{k} \text{SortDescending}(\{l_{\text{task}}(x_i, y_i; \theta) \mid (x_i, y_i) \in \mathcal{D}_g\}) \\
g^* &= \arg\max_{g \in \mathcal{G}} \mathcal{L}_{\text{group}}(g; \theta) \\
\mathcal{L}_{\text{batch}} &= \mathcal{L}_{\text{group}}(g^*; \theta),
\end{align}
where $\mathcal{L}_{\text{group}}(g; \theta)$ represents the loss for group $g$, $k$ is the number of top losses to consider, 
and $g^*$ represents the group with the largest top-k loss. 

For cases where predefined subpopulations aren't available, \citet{broscheit-etal-2022-distributionally} introduce the TopK-AE DRO method where the representations of the encoder of the language model are used to define the latent groups for every sample in the batch. These latent groups are defined by training and updating an autoencoder during the fine-tuning process. The loss function while training the autoencoder consists of a reconstruction loss ($\mathcal{L}_{\text{recon}}$) and a diversity loss ($\mathcal{L}_{\text{div}}$). The reconstruction loss is the cross entropy loss between the reconstructed output and the original input to the autoencoder and the diversity loss is similar to a T-SNE loss~\cite{van2008visualizing} to prevent mode collapse.
The reconstruction losses are stated~\cite{broscheit-etal-2022-distributionally} as follows:
\begin{align}
\mathcal{L}_{\text{recon}}(X_{\text{enc}}, \theta_{\text{AE}}) = -\frac{1}{b}\sum_{i=1}^{b} \log \left( \frac{e^{(X_{\text{enc}} R^T)_{i\cdot}}}{\sum_{k=1}^{d} e^{(X_{\text{enc}} R^T)_{ik}}} \right),
\end{align}
where $R$ is the reconstructed output of the autoencoder for input $X_{enc} \in \mathbb{R}^{bxd}$ with the parameters $\theta_{AE}$, $b$ is the batch size of the autoencoder and $d$ is the hidden size.  The diversity loss is given by:

\begin{align}
    \mathcal{L}_{\text{div}}(X_{enc},\theta_{AE}) = \frac{\sum_{i \neq j} KL(H_i,H_j)}{b(b-1)},
\end{align}
where $KL$ is the Kullback-Leibler divergence applied on the bottleneck layer $H$ of the autoencoder having batch size $b$ and parameters $\theta_{AE}$. $H_i$ refers to the probability distribution output for the $i$-th example from the bottleneck layer $H$.
The overall loss while training the autoencoder can be stated as:
\begin{align}
\mathcal{L}_{\text{AE}} = \mathcal{L}_{\text{recon}} + \beta\mathcal{L}_{\text{div}},
\end{align}
where $\beta$ is a hyperparameter\footnote{We set $\beta=0.1$ as per~\cite{broscheit-etal-2022-distributionally} to regulate the impact of diversity loss relative to reconstruction loss.} and $\mathcal{L}_{\text{AE}}$ is the final loss of the autoencoder.

After defining latent groups for every sentence, the loss is calculated in the same way as the TopK-Group DRO algorithm.
For our experiments, we define the autoencoder to have a batch size $64$, a hidden state size of $128$ with $6$ latent groups.\footnote{These hyperparameters are provided in more detail in Appendix~\ref{app:hyperparams}.}

\section{Proposed Method}
\label{sec:proposed method}

Our proposed method, \textit{RobustDebias}, tackles two primary limitations of existing approaches: (1) most DRO methods require predefined subgroups or identify these groups without bias awareness, and (2) existing debiasing methods can handle only one demographic at a time. We combine the strengths of TopK-Group DRO (using worst-case examples) and TopK-AE DRO (identifying latent groups) while integrating bias demographic information through weighted reconstruction loss. This enables us to debias multiple demographics simultaneously without manually specifying bias-specific word lists.

We replace the ERM objective in the fine-tuning phase with DRO. We formulate DRO in a similar manner to \citet{broscheit-etal-2022-distributionally}, assuming that the training data $D_{train}$ consists of several subpopulations. For our experiments, we consider these subpopulations to be different bias demographics for the Group DRO and TopK-Group DRO approaches. We implement several variants of DRO that differ in how losses across the various subgroups are considered. Through our results discussed in Section~\ref{sec:results}, we observe that using these bias demographics as subpopulations belonging to the uncertainty set $\mathcal{Q}$ for the DRO approach helps introduce a debiasing effect in these models.

Previous work ~\cite{broscheit-etal-2022-distributionally}, has demonstrated that employing DRO for fine-tuning language models on datasets consisting of different distributions of subgroups 
helps reduce spurious correlations and improves performance across all subpopulations. We therefore adopt the DRO approach and designate the bias demographics within the dataset as different subgroups to be optimized.

The proposed approach \textit{RobustDebias}, is divided into two main components. The first is the autoencoder component which takes contextualized representations as input from the encoder model. These representations are then used to train and divide the training data into latent subgroups using the aforementioned objective functions. The second component focuses on the fine-tuning phase of the encoder model. The latent subgroups defined by the autoencoder are used during loss calculation while fine-tuning the model and are defined within the uncertainty set.

The loss calculation follows the TopK-Group DRO method~\cite{broscheit-etal-2022-distributionally}, where losses are initially accumulated per-subgroup and then the top-k largest losses are selected from every subgroup for calculating the final average loss for that batch during fine-tuning of the encoder model.

\textit{RobustDebias} also uses an autoencoder similar to the TopK-AE DRO approach~\cite{broscheit-etal-2022-distributionally}. However, we modify the $\mathcal{L}_{recon}$ loss term of the autoencoder to include information about the bias type that every sample belongs to. This latent information is added by a weight vector representing the inverse frequency of the number of samples of the bias demographic that each sentence in the dataset belongs to.
This helps embed latent information about bias types while creating latent subgroups and thus, helps with the debiasing objective. The modified $\mathcal{L}_{recon}$ term can be stated as:

\begin{align}
    \mathcal{L}_{recon}(X, \theta) = - \frac{1}{b}\sum_{i=1}^{b}\left[\sum_{j=1}^{d}w_{i,j} X_{i,j} \cdot \log \left(\frac{e^{R_{i,j}}}{\sum_{k=1}^{d} e^{R_{k,j}}}\right)\right],
\end{align}

where $R_{i,j}$ is the reconstructed output of the autoencoder for the input $X_{i,j}$ and the weight $w_{i,j}$ is defined as the inverse frequency of the bias type to which the sentence $X_{i,j}$ belongs. We use the $\mathcal{L}_{div}$ term unchanged to prevent mode collapse. The per-sample weights in $\mathcal{L}_{recon}$ also help amplify bias types which are underrepresented in the dataset.

We incorporate a weighted loss to train the autoencoder, as multiple prior works ~\cite{oren2019distributionally,broscheit-etal-2022-distributionally,xie2023doremi} demonstrate that DRO approaches perform poorly with overparameterized models. Furthermore, the suboptimal debiasing performance observed in some existing DRO approaches may arise from imbalances within the bias demographics themselves. This weighted loss for generating latent topics can potentially mitigate this issue. We observe that our approach shows stable performance across multiple models and different dataset sizes.

\section{Experiments}
\label{ch:Experiments}
In this section, we provide details about our experimental setup, including evaluation metrics, the dataset and the three data split configurations for the dataset created to test debiasing performance under different distributional scenarios.

\subsection{Metrics and Testing}
\label{sec:metrics-and-testing}
We evaluate gender, race, and religion biases using three widely-adopted metrics: StereoSet~\cite{nadeem2021stereoset}, SEAT~\cite{may2019measuring}, and CrowS-Pairs~\cite{nangia2020crows}. We selected these due to their popularity in evaluating biases in language models~\cite{meade2022empirical,guo2022auto,zhou-etal-2023-causal,yu-etal-2023-unlearning}.

\paragraph{StereoSet:}
StereoSet~\cite{nadeem2021stereoset} measures bias across gender, profession, race, and religion. The biases are measured on intersentence (masked language modeling) and intrasentence (next sentence prediction) tasks using a Language Modeling Score (LMS) and a Stereotyping Score (SS). For intersentence evaluation, sentences contain blanks that can be filled with stereotypical, anti-stereotypical, or nonsensical terms. Two scores are computed: LMS measures the percentage of responses preferring meaningful terms (stereotype or anti-stereotype) over nonsensical ones, while SS measures the percentage favoring stereotypical over anti-stereotypical completions. An SS of 50\% indicates no bias (random selection between stereotypical and anti-stereotypical). The ICAT (Idealized Context Association Test) score combines both metrics to identify models with strong language modeling capabilities and effective debiasing.

\paragraph{SEAT:}
The Sentence Encoder Association Test (SEAT)~\cite{may2019measuring} extends the word-level WEAT metric~\cite{caliskan2017semantics} to sentence-level evaluation. SEAT measures bias by comparing cosine similarities between contextual embeddings when target words (e.g., "husband," "wife") are swapped in sentences containing attribute words (e.g., "doctor," "nurse"). The effect size is calculated as the difference in these similarities—averaged across multiple tests and indicates bias magnitude, with values closer to 0 representing lower bias. Since encoder-only models produce these contextual embeddings for downstream tasks, mitigating bias during fine-tuning prevents its propagation.  While SEAT only tests the biases in the models, it is a useful metric as it tests some common stereotypes including Double Bind\footnote{Women criticized negatively for both success and failure in professional contexts.} and Angry Black Woman\footnote{Black women stereotyped as rude and angry.} phenomena. The terms bias and stereotype are closely equivalent:  biases may denote implicit terms which differ on a personal level, whereas stereotypes are broader, referring to prejudiced societal beliefs. Both terms have been known to equally influence behaviour towards a particular subgroup~\cite{dovidio2002implicit}.

\paragraph{CrowS-Pairs:}
CrowS-Pairs~\cite{nangia2020crows} provides 1,508 sentence pairs contrasting stereotypes and anti-stereotypes across multiple demographics. A perfectly debiased model assigns equal probability to stereotypical and anti-stereotypical sentences. Unlike StereoSet, this metric focuses solely on bias without assessing language modeling capability. While~\citet{blodgett2021stereotyping} identify limitations in CrowS-Pairs, including dataset noise and sentences that fail to highlight relevant attribute aspects, we include it in our study due to its continued widespread use~\cite{meade2022empirical}.

\subsection{Dataset}
\label{sec:dataset}

We focus on the HolisticBias dataset~\cite{smith-etal-2022-im} which contains 450k sentence templates with 594 descriptor terms across 13 demographic bias categories. For example, ``\texttt{What do you think of middle aged [DESCRIPTOR] dads?}" represents age bias for males, with \texttt{[DESCRIPTOR]} replaced by age-related terms. We use this dataset given its popularity in LLM evaluation~\cite{touvron2023llama,barrault2023seamlessm4t,rastogi2023supporting,guo2023evaluating}. For fine-tuning via MLM, we mask the target word representing the biased subgroup (e.g., \texttt{dads} becomes \texttt{[MASK]}). Compiling the complete dataset with appropriate descriptors\footnote{Instructions from \url{https://github.com/facebookresearch/ResponsibleNLP}} creates approximately 1 million sentences. Since this is a comparatively large dataset relative to other datasets used for fine-tuning language models~\cite{wang2018glue}, we opted to use a smaller subset of this data for fine-tuning and testing the debiasing effect on our language models.

Our evaluation metrics evaluate debiasing effects specifically for gender, race, and religion demographics. Following the approach by~\citet{touvron2023llama}, which evaluates model biases using targeted demographic subsets, we partition the dataset based on the bias demographics present in each sentence. We construct three distinct dataset splits from HolisticBias to examine different distributional configurations:

\subsubsection{3-Bias Split Dataset:}\label{para:3-bias-data}
This split contains only sentences representing the three demographics measured by our metrics (\ref{sec:metrics-and-testing}): gender (53,215 sentences), race (40,737 sentences), and religion (63,124 sentences), totaling 157,076 sentences.

\subsubsection{6-Bias High Split Dataset:}\label{para:6-bias-high-data}
Since~\citet{broscheit-etal-2022-distributionally} demonstrate that DRO variants perform well on minority classes in classification tasks, we construct this split to examine whether similar effects emerge for debiasing. Here, gender, race, and religion function as minority categories alongside the three most-represented demographics: body type (217,631 sentences), age (157,443 sentences), and ability (154,507 sentences). This configuration gives 686,657 sentences.

\subsubsection{6-Bias Low Split Dataset:}\label{para:6-bias-low-data}
To demonstrate that \textit{RobustDebias} improvements extend beyond minority classes, we create a split where gender, race, and religion are majority categories. We augment these with the three least-represented demographics: socioeconomic class (36,333 sentences), nationality (28,993 sentences), and sexual orientation (12,111 sentences), totaling 234,513 sentences.

\section{Results and Discussion}
\label{sec:results}
We now present comprehensive experimental results evaluating RobustDebias against baseline DRO variants and existing debiasing methods. For each method, we fine-tuned the model a single time and store the fine-tuned checkpoint. All reported values in the results correspond to evaluations performed on these stored fine-tuned checkpoints.

\subsection{RobustDebias against baseline DRO variants}
\subsubsection{StereoSet: Measuring Stereotypical Associations}
\label{stereoset}

\begin{table}[!hbt]
\centering
\resizebox{\textwidth}{!}{
\begin{tabular}{ll|ccc|ccc|ccc|ccc}
\hline
& & \multicolumn{3}{c|}{\textbf{BERT}} & \multicolumn{3}{c|}{\textbf{RoBERTa}} & \multicolumn{3}{c|}{\textbf{DistilBERT}} & \multicolumn{3}{c}{\textbf{ALBERT}} \\
\textbf{Split} & \textbf{DRO Type} & \textbf{LMS} & \textbf{SS} & \textbf{ICAT} & \textbf{LMS} & \textbf{SS} & \textbf{ICAT} & \textbf{LMS} & \textbf{SS} & \textbf{ICAT} & \textbf{LMS} & \textbf{SS} & \textbf{ICAT} \\ \hline
\multirow{7}{*}{\rotatebox{90}{3-Bias}}
& ERM & 83.78 & 59.15 & 68.44 & 89.12 & 65.13 & 62.15 & 85.63 & 59.15 & 69.96 & 87.63 & 62.35 & 65.98 \\
& Group & 84.67 & 59.75 & 68.17 & 88.97 & 58.61 & 73.65 & 86.29 & 58.76 & 71.17 & 87.15 & 60.59 & 68.69 \\
& Topic CVaR & 85.13 & 58.61 & 70.47 & 87.63 & 59.63 & 70.75 & 83.48 & 55.42 & 74.43 & 87.69 & 61.65 & 67.25 \\
& TopK & 83.32 & 65.91 & 56.81 & 85.15 & 57.62 & 72.17 & 87.51 & 57.59 & 74.22 & \textbf{88.18} & 57.86 & 74.32 \\
& TopK-Group & \textbf{87.88} & 55.65 & 77.94 & \underline{90.13} & 56.96 & 77.58 & 86.98 & 55.71 & 77.05 & \underline{87.88} & 59.15 & 71.79 \\
& TopK-AE & \underline{85.75} & \underline{53.32} & \underline{80.05} & \textbf{90.27} & \underline{55.18} & \underline{80.92} & \textbf{87.63} & \underline{54.65} & \underline{79.48} & 87.69 & \underline{56.63} & \underline{76.06} \\
& \textit{RobustDebias} & 85.35 & \textbf{52.25} & \textbf{81.51} & 89.95 & \textbf{54.91} & \textbf{81.12} & \underline{87.61} & \textbf{53.78} & \textbf{80.98} & 87.69 & \textbf{55.92} & \textbf{77.31} \\ \hline
\multirow{7}{*}{\rotatebox{90}{6-Bias High}}
& ERM & \underline{85.96} & 63.93 & 62.01 & 90.34 & 66.09 & 61.27 & 86.75 & 65.71 & 59.49 & \textbf{90.63} & 70.15 & 54.11 \\
& Group & 82.65 & 61.44 & 63.74 & 89.51 & 58.90 & 73.57 & 87.52 & 59.19 & 71.43 & 88.51 & 62.34 & 66.67 \\
& Topic CVaR & 83.44 & 64.12 & 59.87 & 87.58 & 60.10 & 69.89 & 85.88 & \underline{54.79} & 77.65 & 88.54 & 62.07 & 67.16 \\
& TopK & 85.55 & 60.86 & 66.96 & 87.34 & 58.51 & 72.47 & 88.16 & 58.01 & 74.04 & 89.27 & 58.91 & 73.36 \\
& TopK-Group & 84.67 & 59.75 & 68.17 & \textbf{91.25} & 57.53 & 77.51 & 87.51 & 57.85 & 73.77 & \underline{90.23} & 60.10 & 72.00 \\
& TopK-AE & 85.66 & \underline{57.44} & \underline{72.91} & 90.39 & \underline{56.81} & \underline{78.08} & \underline{88.45} & 55.97 & \underline{77.89} & 89.62 & \underline{57.83} & \underline{75.59} \\
& \textit{RobustDebias} & \textbf{88.46} & \textbf{56.46} & \textbf{77.03} & \underline{90.76} & \textbf{54.87} & \textbf{81.92} & \textbf{88.91} & \textbf{53.96} & \textbf{81.87} & 88.48 & \textbf{56.21} & \textbf{77.49} \\ \hline
\multirow{7}{*}{\rotatebox{90}{6-Bias Low}}
& ERM & 86.27 & 61.49 & 66.45 & 89.62 & 65.59 & 61.68 & 85.80 & 64.66 & 60.64 & \underline{89.78} & 69.93 & 53.99 \\
& Group & 87.18 & 62.11 & 66.07 & 88.79 & 58.46 & 73.77 & 86.56 & 58.24 & 72.29 & 88.82 & 62.14 & 67.25 \\
& Topic CVaR & 87.66 & 60.93 & 68.50 & 86.88 & 59.65 & 70.11 & 84.94 & \underline{53.91} & 78.30 & 88.10 & 61.87 & 67.19 \\
& TopK & 85.79 & 68.52 & 54.01 & 86.64 & 58.07 & 72.66 & 87.19 & 57.08 & 74.84 & 88.07 & 58.72 & 72.71 \\
& TopK-Group & \textbf{90.49} & 57.85 & 76.28 & 89.67 & 57.10 & 77.67 & 86.55 & 56.92 & 74.57 & \textbf{90.18} & 59.91 & 72.31 \\
& TopK-AE & \underline{88.30} & \underline{55.43} & \underline{78.71} & \textbf{90.52} & \underline{56.38} & \underline{78.23} & \underline{87.48} & 55.07 & \underline{78.61} & 89.17 & \underline{57.64} & \underline{75.54} \\
& \textit{RobustDebias} & 87.89 & \textbf{54.32} & \textbf{80.30} & \underline{90.03} & \textbf{54.46} & \textbf{82.00} & \textbf{87.93} & \textbf{53.10} & \textbf{82.48} & 88.04 & \textbf{56.03} & \textbf{77.42} \\ \hline
\end{tabular}}
\caption{StereoSet evaluation across different model architectures and bias split configurations. LMS indicates Language Modeling Score, SS is the Stereotype Score, ICAT is the overall score. Higher is better for LMS \& ICAT, closer to 50 is better for SS (\textbf{Best}; \underline{Next Best} per model per split)}
\label{tab:stereoset-combined-all}
\end{table}

We evaluate DRO variants on StereoSet~\cite{nadeem2021stereoset} across all models in Table \ref{tab:stereoset-combined-all}. RobustDebias consistently achieves the optimal SS across all model architectures and delivers the best overall ICAT scores among all DRO variants and ERM baselines. For LMS, \textit{RobustDebias} does not always achieve the highest performance. We attribute this to the autoencoder's weighting function, which may reduce performance on certain robust subsets during DRO training. However, the reduction in LMS is relatively minor. In contrast, \textit{RobustDebias} demonstrates a strong debiasing effect. It consistently achieves the optimal SS across all model types, with scores closest to 50 indicating reduced stereotyping. The combined ICAT metric confirms that \textit{RobustDebias} outperforms all other DRO variants and traditional ERM training across all models.

Compared to the 3 bias split, the 6-Bias High configuration shows degraded SS performance but improved LMS across most DRO variants. The larger dataset exhibits improved LMS, while the degraded SS occurs because the three originally targeted demographics (gender, race, religion) become minority samples when three additional bias demographics are introduced. Since StereoSet only evaluates these original three demographics, their minority status in the training data leads to worse stereotyping performance. \textit{RobustDebias} exhibits significantly less performance degradation than other DRO algorithms under this configuration. While DRO methods generally mitigate bias compared to ERM, SS scores still increase substantially for most variants, indicating they don't fully address biases in minority samples. For instance, TopK-AE shows substantial performance drops for BERT between the 3 bias split and 6-Bias High split configurations, whereas \textit{RobustDebias} maintains stable or even improved performance by specifically focusing on minority samples.

Performance in the 6-Bias Low split falls between the 3 bias split and 6-Bias High split configurations. The decline from the 3 bias split occurs for the same reason as previously mentioned: additional demographics are introduced. However, performance improves relative to the 6-Bias High split because gender, race, and religion now constitute majority samples in the training data rather than minority samples. LMS performance worsens compared to the 6-Bias High split, primarily due to the reduced dataset size, though the degradation remains modest. SS improves versus the 6-Bias High split but remains worse than the 3 bias split, as expected given that other bias categories are still present in the data. These results demonstrate that debiasing effectiveness depends on the fine-tuning dataset composition and size. Nevertheless, \textit{RobustDebias} maintains stable performance across all dataset configurations and model architectures.

\subsubsection{SEAT: Measuring Embedding-Level Bias}
\label{seat}

\begin{table}[!hbt]
\centering
\resizebox{\textwidth}{!}{
\begin{tabular}{ll|ccc|ccc|ccc|ccc}
\hline
& & \multicolumn{3}{c|}{\textbf{BERT}} & \multicolumn{3}{c|}{\textbf{RoBERTa}} & \multicolumn{3}{c|}{\textbf{DistilBERT}} & \multicolumn{3}{c}{\textbf{ALBERT}} \\
\textbf{Split} & \textbf{DRO Type} & \textbf{Gender} & \textbf{Race} & \textbf{Religion} & \textbf{Gender} & \textbf{Race} & \textbf{Religion} & \textbf{Gender} & \textbf{Race} & \textbf{Religion} & \textbf{Gender} & \textbf{Race} & \textbf{Religion} \\ \hline
\multirow{7}{*}{\rotatebox{90}{3-Bias}}
& ERM & 0.62 & 0.62 & 0.49 & 0.94 & \textbf{0.14} & \underline{0.35} & 0.72 & 0.67 & 0.60 & 0.69 & 0.61 & 0.58 \\
& Group & 0.57 & 0.59 & 0.38 & 0.88 & 0.25 & 0.38 & 0.47 & \textbf{0.53} & 0.45 & 0.49 & 0.45 & 0.49 \\
& Topic CVaR & 0.51 & 0.64 & 0.43 & 0.91 & 0.39 & 0.43 & 0.56 & 0.59 & 0.48 & 0.61 & 0.43 & 0.57 \\
& TopK & 0.65 & 0.59 & 0.45 & 0.76 & 0.31 & 0.45 & 0.54 & 0.62 & 0.71 & \textbf{0.35} & \underline{0.42} & 0.51 \\
& TopK-Group & 0.35 & 0.41 & 0.40 & 0.61 & 0.18 & 0.40 & 0.53 & \underline{0.55} & 0.49 & 0.39 & 0.47 & 0.59 \\
& TopK-AE & \underline{0.27} & \textbf{0.39} & \underline{0.31} & \textbf{0.45} & \underline{0.17} & \textbf{0.32} & \underline{0.46} & \textbf{0.53} & \underline{0.42} & 0.39 & 0.45 & \underline{0.32} \\
& \textit{RobustDebias} & \textbf{0.25} & \underline{0.40} & \textbf{0.22} & \underline{0.49} & \textbf{0.14} & \textbf{0.32} & \textbf{0.44} & \textbf{0.53} & \textbf{0.39} & \underline{0.38} & \textbf{0.41} & \textbf{0.30} \\ \hline
\multirow{7}{*}{\rotatebox{90}{6-Bias High}}
& ERM & 0.84 & 0.76 & 0.61 & 0.97 & 0.54 & 0.69 & 0.89 & 0.72 & 0.72 & 0.75 & 0.87 & 0.71 \\
& Group & \textbf{0.43} & 0.56 & 0.52 & 0.91 & 0.37 & 0.76 & \underline{0.51} & 0.68 & 0.61 & 0.54 & \textbf{0.47} & 0.65 \\
& Topic CVaR & 0.60 & 0.66 & 0.72 & 0.94 & 0.35 & 0.59 & 0.85 & 0.63 & \underline{0.52} & 0.77 & 0.84 & 0.63 \\
& TopK & 0.59 & 0.75 & \underline{0.45} & \textbf{0.41} & 0.66 & 0.55 & 0.73 & 0.87 & 0.88 & 0.68 & \underline{0.49} & 0.76 \\
& TopK-Group & \underline{0.58} & \textbf{0.45} & 0.61 & 0.70 & \underline{0.37} & 0.57 & 0.59 & \underline{0.65} & 0.76 & 0.61 & 0.64 & 0.73 \\
& TopK-AE & 0.60 & 0.59 & 0.65 & 0.61 & 0.49 & \underline{0.46} & 0.78 & 0.71 & 0.66 & \underline{0.47} & 0.78 & \underline{0.57} \\
& \textit{RobustDebias} & \underline{0.58} & \underline{0.50} & \textbf{0.38} & \underline{0.55} & \textbf{0.21} & \textbf{0.39} & \textbf{0.49} & \textbf{0.57} & \textbf{0.39} & \textbf{0.43} & \textbf{0.47} & \textbf{0.35} \\ \hline
\multirow{7}{*}{\rotatebox{90}{6-Bias Low}}
& ERM & 0.65 & 0.65 & 0.51 & 0.93 & 0.47 & 0.65 & 0.86 & 0.68 & 0.67 & 0.73 & 0.85 & 0.69 \\
& Group & 0.59 & 0.62 & 0.40 & 0.88 & 0.35 & 0.53 & 0.49 & 0.65 & 0.57 & 0.53 & \underline{0.55} & 0.63 \\
& Topic CVaR & 0.53 & 0.67 & 0.45 & 0.67 & 0.33 & 0.55 & 0.82 & \underline{0.60} & \underline{0.49} & 0.75 & 0.82 & 0.61 \\
& TopK & 0.68 & 0.62 & 0.47 & 0.59 & 0.47 & 0.51 & 0.70 & 0.83 & 0.82 & \textbf{0.42} & 0.48 & 0.74 \\
& TopK-Group & 0.37 & 0.43 & 0.42 & 0.91 & \underline{0.35} & 0.71 & \underline{0.57} & 0.62 & 0.71 & 0.60 & 0.62 & 0.71 \\
& TopK-AE & \underline{0.28} & \textbf{0.41} & \underline{0.32} & \textbf{0.39} & 0.63 & \underline{0.43} & 0.75 & 0.67 & 0.62 & \underline{0.46} & 0.76 & \underline{0.55} \\
& \textit{RobustDebias} & \textbf{0.26} & \underline{0.42} & \textbf{0.23} & \underline{0.53} & \textbf{0.20} & \textbf{0.37} & \textbf{0.47} & \textbf{0.54} & \textbf{0.37} & 0.67 & \textbf{0.46} & \textbf{0.34} \\ \hline
\end{tabular}}
\caption{SEAT: Average effect size for each demographic across different model architectures and bias split configurations. Lower is better (\textbf{Best}; \underline{Next Best} per model per split)}
\label{tab:seat-combined-all}
\end{table}

Table \ref{tab:seat-combined-all} shows the performance of the given DRO variants for different model types on the SEAT metric~\cite{may2019measuring}. Similar to the StereoSet metric, we observe that overall \textit{RobustDebias} exhibits a strong debiasing effect when compared to other DRO variants for all models, achieving the best or second-best performance for all models across all the bias demographics it was evaluated on. While \textit{RobustDebias} does not achieve the best performance for all demographics, it should be noted that the loss in performance may also be due to the intricate nature of stereotypes studied under the SEAT metric, as the authors of ~\cite{may2019measuring} proposed specific word lists and it is possible that \textit{RobustDebias} does not perform well on some of these word lists. The loss in performance is quite negligible.

For the 6-Bias High split, an erratic trend of performances by TopK DRO and TopK-AE DRO approaches is observed for the RoBERTa model. TopK performs well with the best score for the gender demographic, but its performance degrades for the other demographics, which may be due to TopK-DRO not being able to handle losses of all subgroups since the subgroups aren't defined for the TopK-DRO approach. Comparing other DRO variants with \textit{RobustDebias} reveals very stable performance across all demographics, and it should be noted that unlike \textit{RobustDebias}, none of the other prominent debiasing methods debias multiple bias demographics in the same model while performing the best for all the metrics being evaluated.

For the 6-Bias High split, an erratic trend of performance by TopK DRO and TopK-AE DRO approaches is observed for the RoBERTa model. TopK performs well with the best score for the gender demographic, but its performance deteriorates for the other demographics, which may be due to TopK-DRO being unable to handle losses of all subgroups since the subgroups aren't defined for the TopK-DRO approach. Comparing other DRO variants with \textit{RobustDebias} reveals highly stable performance across all demographics. \textit{RobustDebias} achieves strong performance across multiple bias demographics simultaneously within a single model, whereas other prominent debiasing methods fail to maintain such consistency across all the metrics being evaluated.

For the 6-Bias Low split, similar trends emerge as with the StereoSet metric when comparing the results on SEAT with the 3 bias split experiments and the 6 bias high split experiments. While TopK-AE also performs strongly, obtaining the best and second-best scores for almost all demographics across BERT and RoBERTa models, there is a substantial degradation in performance for DistilBERT and ALBERT models. The TopK-AE DRO approach seems unpredictable in terms of debiasing performance as it depends heavily on the latent topics which are trained without any specific information about bias demographics.

\subsubsection{CrowS-Pairs: Measuring Stereotypical Probability Assignments}
\label{}

\begin{table}[!hbt]
\centering
\resizebox{\textwidth}{!}{
\begin{tabular}{ll|ccc|ccc|ccc|ccc}
\hline
& & \multicolumn{3}{c|}{\textbf{BERT}} & \multicolumn{3}{c|}{\textbf{RoBERTa}} & \multicolumn{3}{c|}{\textbf{DistilBERT}} & \multicolumn{3}{c}{\textbf{ALBERT}} \\
\textbf{Split} & \textbf{DRO Type} & \textbf{Gender} & \textbf{Race} & \textbf{Religion} & \textbf{Gender} & \textbf{Race} & \textbf{Religion} & \textbf{Gender} & \textbf{Race} & \textbf{Religion} & \textbf{Gender} & \textbf{Race} & \textbf{Religion} \\ \hline
\multirow{7}{*}{\rotatebox{90}{3-Bias}}
& ERM & 57.65 & 62.39 & 64.95 & 60.32 & 65.78 & 59.98 & 40.29 & 43.19 & 67.85 & 40.15 & 69.88 & 59.79 \\
& Group & 57.25 & 62.33 & 62.86 & 55.75 & 60.91 & 58.86 & 43.11 & 39.85 & 60.11 & 43.97 & 55.14 & 43.65 \\
& Topic CVaR & 59.51 & 60.52 & 62.02 & 63.95 & 57.15 & 69.85 & 39.67 & 64.12 & 62.02 & 39.85 & 25.17 & 36.79 \\
& TopK & \underline{53.13} & 58.51 & 65.31 & 58.26 & 61.22 & 57.35 & 40.51 & 42.69 & 68.83 & \textbf{46.65} & 45.22 & \underline{46.35} \\
& TopK-Group & 55.28 & 57.12 & 54.20 & 55.50 & 56.17 & 55.29 & 39.71 & 34.42 & 62.65 & 44.18 & 45.61 & 40.91 \\
& TopK-AE & 53.14 & \textbf{54.23} & \underline{52.22} & \underline{54.45} & \underline{55.09} & \underline{52.59} & \textbf{46.32} & \underline{45.61} & \underline{58.12} & 45.98 & \underline{53.65} & 57.91 \\
& \textit{RobustDebias} & \textbf{52.25} & \underline{56.67} & \textbf{50.89} & \textbf{53.12} & \textbf{54.33} & \textbf{52.11} & \underline{44.85} & \textbf{45.98} & \textbf{55.67} & \underline{53.49} & \textbf{52.11} & \textbf{53.24} \\ \hline
\multirow{7}{*}{\rotatebox{90}{6-Bias High}}
& ERM & 62.56 & 68.16 & 66.67 & 62.57 & 68.89 & 66.78 & 36.71 & 65.19 & 63.40 & 67.85 & 35.33 & 32.19 \\
& Group & 60.55 & 62.33 & 40.89 & \underline{54.89} & 61.34 & 59.65 & 58.23 & 62.51 & \underline{58.76} & 40.91 & 57.88 & 39.51 \\
& Topic CVaR & 66.74 & \underline{61.29} & 70.54 & 67.99 & 64.81 & 72.35 & 25.14 & 39.67 & 40.88 & 63.44 & 66.90 & 69.91 \\
& TopK & 55.98 & 63.99 & \textbf{56.65} & 56.69 & 63.47 & 57.14 & \textbf{46.87} & \underline{43.44} & 59.61 & 39.79 & 59.84 & 40.12 \\
& TopK-Group & \textbf{53.78} & 68.15 & 64.45 & 57.78 & 57.63 & 58.36 & 39.83 & 62.84 & 35.56 & 39.28 & 37.79 & \underline{59.16} \\
& TopK-AE & 57.25 & 63.39 & 65.78 & 57.81 & \underline{56.65} & \underline{55.13} & 57.61 & 43.11 & 60.42 & \textbf{44.87} & \underline{56.69} & 60.12 \\
& \textit{RobustDebias} & \underline{55.56} & \textbf{59.73} & \underline{62.86} & \textbf{54.45} & \textbf{54.76} & \textbf{52.09} & \underline{44.97} & \textbf{46.87} & \textbf{56.24} & \textbf{55.13} & \textbf{47.65} & \textbf{45.79} \\ \hline
\multirow{7}{*}{\rotatebox{90}{6-Bias Low}}
& ERM & 59.35 & 65.88 & 67.03 & 62.44 & 67.93 & 65.04 & 36.49 & 64.47 & 61.05 & 65.20 & 34.27 & 31.80 \\
& Group & 58.94 & 65.82 & 64.87 & \underline{54.78} & 60.48 & 58.10 & 57.88 & 61.82 & \underline{56.59} & 39.31 & 56.14 & 39.04 \\
& Topic CVaR & 61.27 & 63.91 & 64.00 & 67.85 & 63.90 & 70.47 & 24.99 & 39.23 & 39.37 & 60.97 & 64.89 & 69.07 \\
& TopK & \underline{54.70} & 61.79 & 67.40 & 56.58 & 62.58 & 55.65 & \underline{44.70} & \underline{42.96} & 34.24 & 38.24 & 58.04 & 39.64 \\
& TopK-Group & 56.91 & 60.32 & 55.93 & 57.66 & 56.82 & 56.84 & 39.59 & 62.15 & 57.40 & 37.75 & 36.66 & \underline{58.45} \\
& TopK-AE & 54.71 & \textbf{57.27} & \underline{53.89} & 57.69 & \underline{55.86} & \underline{53.70} & 57.26 & 42.64 & 58.18 & \underline{43.12} & \underline{53.99} & 59.40 \\
& \textit{RobustDebias} & \textbf{53.79} & \underline{59.84} & \textbf{52.52} & \textbf{54.34} & \textbf{53.99} & \textbf{53.61} & \textbf{46.59} & \textbf{46.35} & \textbf{54.16} & \textbf{52.98} & \textbf{46.22} & \textbf{45.24} \\ \hline
\end{tabular}}
\caption{CrowS-Pairs: Metric scores for each demographic across different model architectures and bias split configurations. Closer to 50 is better (\textbf{Best}; \underline{Next Best} per model per split)}
\label{tab:crows-combined-all}
\end{table}

We present the results for the DRO variants across all models on the CrowS-Pairs evaluation metric~\cite{nangia2020crows} in Table \ref{tab:crows-combined-all}. Similar to the SEAT metric, we observe that \textit{RobustDebias} performs strongly for all models. \textit{RobustDebias} achieves the best or second-best debiasing score for all demographics irrespective of the model type and the data split.

For the 3-Bias split, Topic CVaR shows poor performance across all models and metrics due to the ineffectiveness of using LDA for subset creation in larger datasets, where the resulting topics fail to relate to biases. A similar performance degradation was also observed by ~\cite{broscheit-etal-2022-distributionally} in their classification tasks, demonstrating that DRO approaches alone are not inherently debiasing without bias-specific information. TopK-AE shows strong but inconsistent performance across different metrics and models because the autoencoder forms latent topics without any metadata about bias demographics, whereas \textit{RobustDebias} outperforms these DRO variants across all models and metrics by incorporating such information.

For the 6-Bias High split, for the BERT model specifically, TopK-Group DRO performs best on the gender demographic, but this performance proves unstable across other models, appearing only for specific gender demographics in other metrics. While TopK-Group DRO defines subgroups precisely by bias demographics, it fails to capture certain nuances within those subgroups, an issue that \textit{RobustDebias} addresses by forming latent subgroups using an autoencoder with bias demographic information as metadata. Compared to the 3 bias split dataset, the performance gap between TopK-AE and \textit{RobustDebias} widens substantially, a pattern that also appears in StereoSet and SEAT metrics, with TopK-AE's degraded performance reflecting its inability to mitigate biases in minority samples for larger datasets since its latent groups are formed without considering all bias demographics.

For the 6-Bias Low split, \textit{RobustDebias} shows minimal performance loss when comparing this dataset with the 3 bias split dataset. This stability demonstrates that additional bias categories do not adversely affect \textit{RobustDebias}'s debiasing process.

Across all metrics, we observe that there is a slight improvement in debiasing performance when using a smaller dataset rather than a larger one. However, we do observe that when comparing performance across all metrics for all models, the DRO approaches vary considerably in how they perform on different datasets. This is primarily due to the lack of an objective function which specifically determines the subgroups based on bias demographics. This comes with the exception of \textbf{\textit{RobustDebias}, which performs in a robust manner for debiasing these models for several bias demographics simultaneously.}

\subsection{Comparison with other Debiasing Methods}
\label{debiasing-comp}
We compare the results of the \textit{RobustDebias} approach to other debiasing methods such as AutoDebias~\cite{guo2022auto}, CausalDebias~\cite{zhou-etal-2023-causal}, and PCGU~\cite{yu-etal-2023-unlearning}, and FineDeb~\cite{saravanan2023finedeb}. For this experiment, we use the \textit{RobustDebias} BERT model from the 3 Bias Type Split experiment. We only report the results for the gender and race demographics for SEAT and CrowS-Pairs metrics as the AutoDebias~\cite{guo2022auto} and CausalDebias~\cite{zhou-etal-2023-causal} methods are designed to work exclusively with these bias demographics.

\begin{table*}[!hbt]
\centering
\begin{tabular}{l|ccc|cc|cc}
\hline
& \multicolumn{3}{c|}{\textbf{StereoSet}} & \multicolumn{2}{c|}{\textbf{SEAT}} & \multicolumn{2}{c}{\textbf{CrowS-Pairs}} \\
\textbf{Debiasing Method} & \textbf{LMS} & \textbf{SS} & \textbf{ICAT} & \textbf{Gender} & \textbf{Race} & \textbf{Gender} & \textbf{Race} \\ \hline
AutoDebias & 63.71 & 53.16 & 59.68 & 0.31 & 0.77 & \underline{46.85} & \textbf{47.51} \\
CausalDebias & 74.85 & \underline{52.91} & 70.49 & 0.38 & 0.64 & 46.75 & \underline{52.89} \\
PCGU & 84.71 & 53.75 & \underline{78.36} & \textbf{0.19} & 0.79 & 54.80 & 54.77 \\
FineDeb & \underline{85.23} & 54.65 & 77.30 & 0.39 & \underline{0.51} & 54.55 & 53.69 \\
\textit{RobustDebias (ours)} & \textbf{85.35} & \textbf{52.25} & \textbf{81.51} & \underline{0.25} & \textbf{0.40} & \textbf{52.25} & 56.67 \\ \hline
\end{tabular}
\caption{Comparison of different debiasing methods for BERT across three evaluation metrics. StereoSet: Higher is better for LMS \& ICAT, closer to 50 is better for SS. SEAT: Lower is better. CrowS-Pairs: Closer to 50 is better. (\textbf{Best}; \underline{Next Best})}
\label{tab:bert-debiasing-comparison}
\end{table*}

Table \ref{tab:bert-debiasing-comparison} give the results for comparing some prominent debiasing methods with our \textit{RobustDebias} approach. We observe that our method \textit{RobustDebias} outperforms all the other debiasing methods by giving the best and second-best scores for almost all metrics.

For the StereoSet metric, we observe that \textit{RobustDebias} performs the best in LMS measure as well as the SS measure. Moreover, we also observe that the AutoDebias method which shows a very strong debiasing effect, degrades the LMS measure severely. As per the StereoSet metric, PCGU also seems to perform quite well after \textit{RobustDebias}.

For the SEAT metric and the CrowS-Pairs metric, we observe that \textit{RobustDebias} performs the best or second-best for gender as well as race demographics. AutoDebias performs the best for the race demographic of the CrowS-Pairs metrics. However, it should be noted here that the AutoDebias, CausalDebias and FineDeb methods do not train a single model for debiasing across different demographics. Instead, specific models for each bias demographic are trained separately. Meanwhile, the \textbf{\textit{RobustDebias} approach trains a single model debiased across $3$ bias demographics and still performs better or similar to these debiasing methods.}

\section{Conclusion}
\label{ch:Conclusion}

We have presented a novel approach of debiasing language models using DRO where the optimization itself performs debiasing. Through experimentation and analysis, our modified DRO approach, \textit{RobustDebias} shows stable performance across all models and dataset types when compared to ERM and other DRO variants. 

While the LMS measure is not always optimal for \textit{RobustDebias}, its combined performance across all metrics shows \textit{RobustDebias} is a highly stable debiasing approach across all models and dataset types. Although some DRO variants occasionally outperform \textit{RobustDebias} on the LMS measure, \textit{RobustDebias} consistently achieves the best combined ICAT measure. While not always optimal for LMS, it remains close to the best while demonstrating superior debiasing ability. 

Superior performance on widely used benchmarks does not guarantee the best model. Models trained using ERM perform well on standard benchmarks\footnote{The leaderboard is available at \url{https://gluebenchmark.com/leaderboard}}, yet show substantially higher stereotype scores across all bias metrics compared to DRO. A model performing better on a big benchmark could also be due to some shortcut learning behaviour~\cite{broscheit-etal-2022-distributionally} rather than the model actually being robust and well-trained. Such models are usually skewed towards the majority classes.

Future directions for expanding this work include establishing stronger bias metrics that identify the amount of debiasing in said models without extensive demographic categorization. Substituting fine-tuning tasks with prompt-based tasks such as Multiple Choice Question Answering would reveal debiasing effectiveness when choices are limited rather than examining the entire vocabulary. Moreover, collecting data for other gender demographic classes and training models for multiclass gender debiasing would be valuable. Expanding to different modalities would highlight gaps in prominent models across different modalities. While debiasing metrics may differ significantly across modalities, they would provide insights into DRO's debiasing effects beyond text. 

\section{End Matter}
\subsection{Ethical considerations statement}
To encourage transparency in research, we disclose ethical limitations of our work in the hope that it offers further insights into our proposed framework. While our approach effectively reduces bias across various demographics according to existing metrics, the foundational pretrained models could still be leveraged to recognize personal information that may have been memorized during their original pretraining data. This work focuses on gender debiasing in a binary setting due to the lack of previous work in gender as a multiclass setting. However, we acknowledge this is a significant limitation. Non-binary, genderqueer, and genderfluid individuals are erased by binary framings of gender. The parameters we use to assess stereotype scores and measure the effectiveness of debiasing techniques are commonly used, yet they may not comprehensively capture the true extent of debiasing achieved within the model. The present metrics for bias primarily hinge on the comparison of either target words or attribute words, leading to inconsistent performance across diverse techniques. This indicates the necessity for a more inclusive bias metric that remains indifferent to the specific debiasing technique used. This work deals with the modality of biases to be only textual in nature. However, with the advent of new multimodal models, it is essential to look at biases in other modalities as well as subtler contexts.

\subsection{Generative AI usage statement}
For this paper, LLM tools (Claude Sonnet 4.5 and ChatGPT GPT 5.2) were only used to format equations, tables, improve language clarity and grammar. No original content was generated by AI models. 

\bibliographystyle{unsrt}
\bibliography{sample-base}

\appendix
\section{Appendix}
\subsection{Hyperparameters}\label{app:hyperparams}
For training our language model we use a batch size of $32$ to maximise GPU usage. For language model fine-tuning, we find $20$ epochs to be sufficient, with the training process utilizing an early stopping mechanism. We use a learning rate of $1e-4$ based on what we find in the original BERT paper \cite{devlin2019bert}. Similarly, we use a learning rate of $2e-5$ for RoBERTa as described in the original paper~\cite{liu2019roberta} and a learning rate of $1e-4$ for DistilBERT~\cite{sanh2019distilbert} and ALBERT~\cite{lan2019albert}.

For the DRO mechanisms, we find the $\alpha$ percentile value of $0.8$ to be the best from different hyperparameter sweeps. For the Group DRO setting, we use the default hyperparameters used in the original paper~\cite{sagawa2019distributionally}. For our experiments, we define the autoencoder to have a batch size $64$, a hidden state size of $128$. For the Topic CVaR DRO, we use a chunk size of $4000$ and train the LDA model with $350$ iterations. We find that for the TopK-DRO, TopK-Group DRO, TopK-AE DRO and \textit{RobustDebias}, we find the value of $k=6$ to be the best. We report the results as an average of $5$ seeds picked at random.

\subsection{Training Details}
For each model fine-tuning we run each job with $16$ CPUs per job, $40GB$ memory and 1 NVIDIA V100 GPU. We implement all the models using Huggingface libraries~\cite{wolf2020transformers} and PyTorch~\cite{paszke-pytorch-2019}.

\subsection{Phased Fine-tuning on 6 bias split datasets}

\begin{table}
\centering
\resizebox{0.8\textwidth}{!}{
\begin{tabular}{llccc|ccc}
\hline
\multicolumn{2}{l}{} & \multicolumn{3}{c|}{\textbf{High Split}} & \multicolumn{3}{c}{\textbf{Low Split}}\\\hline
\multicolumn{1}{l}{\textbf{Model}} & \multicolumn{1}{l}{\textbf{DRO Type}} & \textbf{LMS} & \textbf{SS} & \textbf{ICAT} & \textbf{LMS} & \textbf{SS} & \textbf{ICAT}\\ \hline
\multirow{7}{*}{BERT}
& ERM & 85.96 & 64.15 & 61.63 & 86.27 & 60.55 & 68.07 \\
& Group & 83.78 & 60.06 & 66.92 & 87.18 & 63.16 & 64.23 \\
& Topic CVaR & 83.44 & 64.09 & 59.93 & 87.66 & 59.95 & 70.22 \\
& TopK & 85.97 & 61.53 & 66.15 & 85.79 & 68.14 & 54.67 \\
& TopK-Group & 84.67 & 60.55 & 66.80 & \textbf{90.49} & 57.85 & 76.28 \\
& TopK-AE & \underline{86.77} & \underline{57.12} & \underline{74.41} & \underline{88.30} & \underline{56.19} & \underline{77.37} \\
& \textit{RobustDebias} & \textbf{88.46} & \textbf{56.51} & \textbf{76.94} & 87.89 & \textbf{55.97} & \textbf{77.40} \\
\hline
\multirow{7}{*}{RoBERTa}
& ERM & 90.34 & 67.98 & 57.85 & 88.51 & 67.90 & 56.82 \\
& Group & 89.51 & 59.65 & 72.23 & 88.79 & 58.46 & 73.77 \\
& Topic CVaR & 87.58 & 60.10 & 69.89 & 86.88 & 59.69 & 70.04 \\
& TopK & 87.34 & 58.51 & 72.47 & 86.64 & 58.10 & 72.6 \\
& TopK-Group & 90.36 & 58.62 & 74.78 & 89.67 & 57.60 & 76.04 \\
& TopK-AE & \underline{90.39} & \underline{55.14} & \underline{81.10} & \textbf{90.52} & \underline{56.38} & \underline{78.97} \\
& \textit{RobustDebias} & \textbf{90.41} & \textbf{54.97} & \textbf{81.42} & \underline{90.03} & \textbf{55.09} & \textbf{80.86} \\
\hline
\multirow{7}{*}{DistilBERT}
& ERM & 86.75 & 65.71 & 59.49 & 85.80 & 63.09 & 63.34 \\
& Group & 87.52 & 59.19 & 71.43 & 86.56 & 55.31 & 77.37 \\
& Topic CVaR & 85.88 & 54.79 & 77.65 & 84.94 & 53.91 & 78.3 \\
& TopK & 89.13 & 59.63 & 71.96 & 87.19 & 56.65 & 75.59 \\
& TopK-Group & 87.51 & 57.85 & 73.77 & 86.55 & 56.19 & 75.84 \\
& TopK-AE & \underline{87.63} & \underline{54.85} & \underline{79.13} & \underline{87.48} & \underline{55.11} & \underline{78.54} \\
& \textit{RobustDebias} & \textbf{87.98} & \textbf{54.05} & \textbf{80.85} & \textbf{87.65} & \textbf{53.05} & \textbf{82.30} \\
\hline
\multirow{7}{*}{ALBERT}
& ERM & 89.83 & 71.21 & 51.72 & 89.01 & 68.90 & 55.36 \\
& Group & 88.51 & 63.01 & 65.48 & 88.82 & 63.42 & 64.98 \\
& Topic CVaR & 88.67 & 62.07 & 67.27 & 89.34 & 60.95 & 69.77 \\
& TopK & 89.27 & 57.67 & 75.58 & 88.16 & 58.63 & 72.94 \\
& TopK-Group & \textbf{90.34} & 58.99 & 74.10 & \textbf{90.00} & 60.01 & 71.98 \\
& TopK-AE & \underline{89.79} & \underline{56.97} & \underline{77.27} & \underline{89.98} & \underline{58.51} & \underline{74.67} \\
& \textit{RobustDebias} & 89.01 & \textbf{54.34} & \textbf{81.28} & 89.54 & \textbf{57.15} & \textbf{76.74} \\
\hline
\end{tabular}}
\caption{StereoSet evaluation across all models with phased fine-tuned models. LMS indicates Language Modeling Score, SS is the Stereotype Score, ICAT is the overall score. Higher is better for LMS \& ICAT, closer to 50 is better for SS (\textbf{Best}; \underline{Next Best})}
\label{fig:stereoset-all-models-6bph}
\end{table}

We conduct a phased fine-tuning experiment where we divide the fine-tuning of these models into different phases. For the first phase, we fine-tune the models with the 3 bias dataset only. After this fine-tuning, we use these models and fine-tune them again on a dataset of other $3$ bias demographics. These biases are the additional bias demographics in either the 6-bias high split dataset or the 6-bias low split dataset.

We present the results for this phased fine-tuning on the StereoSet metric, the SEAT metric and the CrowS-Pairs metric. 

\begin{table}
\small
\centering
\resizebox{0.8\textwidth}{!}{
\begin{tabular}{llccc|ccc}
\hline
\multicolumn{2}{l}{} & \multicolumn{3}{c|}{\textbf{High Split}} & \multicolumn{3}{c}{\textbf{Low Split}}\\\hline
\multicolumn{1}{l}{\textbf{Model}} & \multicolumn{1}{l}{\textbf{DRO Type}} & \multicolumn{1}{c}{\textbf{Gender}} & \multicolumn{1}{c}{\textbf{Race}} & \multicolumn{1}{c|}{\textbf{Religion}} & \multicolumn{1}{c}{\textbf{Gender}} & \multicolumn{1}{c}{\textbf{Race}} & \multicolumn{1}{c}{\textbf{Religion}} \\ \hline
\multirow{7}{*}{BERT}
& ERM & 0.89 & 0.79 & 0.62 & 0.66 & 0.64 & 0.59 \\
& Group & \textbf{0.47} & 0.56 & 0.52 & 0.59 & 0.61 & 0.41 \\
& Topic CVaR & 0.69 & 0.66 & 0.75 & 0.51 & 0.63 & 0.43 \\
& TopK & 0.57 & 0.75 & \underline{0.45} & 0.71 & 0.55 & 0.47 \\
& TopK-Group & 0.58 & \textbf{0.44} & 0.61 & \underline{0.33} & \underline{0.43} & 0.43 \\
& TopK-AE & 0.60 & 0.59 & 0.67 & \textbf{0.26} & \textbf{0.41} & \underline{0.39} \\
& \textit{RobustDebias} & \underline{0.55} & \underline{0.50} & \textbf{0.39} & \textbf{0.26} & \textbf{0.41} & \textbf{0.27} \\
\hline
\multirow{7}{*}{RoBERTa}
& ERM & 0.99 & 0.54 & 0.69 & 0.95 & 0.47 & 0.65 \\
& Group & 0.93 & 0.39 & 0.76 & 0.88 & 0.35 & 0.53 \\
& Topic CVaR & 0.94 & \underline{0.35} & 0.61 & 0.67 & \underline{0.33} & 0.55 \\
& TopK & \textbf{0.44} & 0.67 & \underline{0.55} & 0.60 & 0.48 & \underline{0.51} \\
& TopK-Group & 0.72 & 0.37 & 0.57 & 0.91 & 0.35 & 0.73 \\
& TopK-AE & 0.63 & 0.51 & 0.59 & \textbf{0.39} & 0.65 & 0.43 \\
& \textit{RobustDebias} & \underline{0.59} & \textbf{0.25} & \textbf{0.40} & \underline{0.54} & \textbf{0.25} & \textbf{0.39} \\
\hline
\multirow{7}{*}{DistilBERT}
& ERM & 0.96 & 0.77 & 0.72 & 0.92 & 0.76 & 0.79 \\
& Group & \textbf{0.51} & 0.68 & \underline{0.62} & \underline{0.49} & 0.65 & 0.57 \\
& Topic CVaR & 0.86 & 0.68 & 0.57 & 0.82 & \underline{0.60} & \underline{0.49} \\
& TopK & 0.73 & 0.87 & 0.88 & 0.69 & 0.83 & 0.82 \\
& TopK-Group & \underline{0.63} & \underline{0.67} & 0.76 & 0.57 & 0.72 & 0.71 \\
& TopK-AE & 0.79 & 0.71 & 0.66 & 0.75 & 0.67 & 0.71 \\
& \textit{RobustDebias} & \textbf{0.51} & \textbf{0.57} & \textbf{0.39} & \textbf{0.47} & \textbf{0.54} & \textbf{0.37} \\
\hline
\multirow{7}{*}{ALBERT}
& ERM & 0.79 & 0.87 & 0.71 & 0.73 & 0.85 & 0.69 \\
& Group & \underline{0.54} & \textbf{0.47} & 0.65 & 0.53 & 0.59 & 0.63 \\
& Topic CVaR & 0.77 & 0.97 & \underline{0.63} & 0.75 & 0.83 & 0.62 \\
& TopK & 0.68 & \underline{0.61} & 0.82 & \textbf{0.42} & \underline{0.47} & 0.77 \\
& TopK-Group & 0.61 & 0.64 & 0.73 & 0.60 & 0.62 & 0.71 \\
& TopK-AE & 0.43 & 0.65 & 0.65 & \underline{0.46} & 0.76 & \underline{0.55} \\
& \textit{RobustDebias} & \textbf{0.43} & \textbf{0.47} & \textbf{0.35} & 0.54 & \textbf{0.46} & \textbf{0.34} \\
\hline
\end{tabular}}
\caption{SEAT: Average effect size for each demographic across all models with phased finetuned split models. Lower is better (\textbf{Best}; \underline{Next Best})}
\label{table:seat-all-models-6bph}
\end{table}
On comparing with the combined experiments, we observe that there is no considerable difference in performance of the models on all DRO variants. For the StereoSet metric, we observe that the LMS measure doesn't change much for both the high split and low split experiments. This is because the overall data that the model is being fine-tuned on doesn't change and thus the performance for the StereoSet metric doesn't vary much. We can also observe this trend for the SS measure where the change in scores is pretty minimal.

RobustDebias approach remains stable under this condition showing that the debiasing effect is not highly degraded by the phased fine-tuning. On comparing with the 3 bias split experiment, we observe that the RobustDebias approach performs in a stable manner and doesn't degrade much in terms of bias scores across CrowS-Pairs, SEAT as well as the StereoSet metric. This shows the robust nature of the approach where the debiasing effect remains steady even when fine-tuning the model with datasets much bigger (high-split dataset) than the original dataset it is fine-tuned on.
\begin{table}
\centering
\resizebox{0.8\textwidth}{!}{
\begin{tabular}{llccc|ccc}
\hline
\multicolumn{2}{l}{} & \multicolumn{3}{c|}{\textbf{High Split}} & \multicolumn{3}{c}{\textbf{Low Split}}\\\hline
\multicolumn{1}{l}{\textbf{Model}} & \multicolumn{1}{l}{\textbf{DRO Type}} & \multicolumn{1}{c}{\textbf{Gender}} & \multicolumn{1}{c}{\textbf{Race}} & \multicolumn{1}{c|}{\textbf{Religion}} & \multicolumn{1}{c}{\textbf{Gender}} & \multicolumn{1}{c}{\textbf{Race}} & \multicolumn{1}{c}{\textbf{Religion}} \\ \hline
\multirow{7}{*}{BERT} 
& ERM & 63.78 & 66.48 & 65.52 & 60.77 & 64.90 & 66.96 \\
& Group & 62.79 & 63.99 & \underline{40.89} & 58.94 & 65.82 & 64.87 \\
& Topic CVaR & 65.79 & 62.75 & 69.50 & 62.25 & 63.91 & 64.12 \\
& TopK & \underline{55.98} & 63.99 & \textbf{56.65} & 54.70 & 62.79 & 67.40 \\
& TopK-Group & \textbf{53.78} & 67.56 & 64.45 & 56.91 & 60.32 & 55.93 \\
& TopK-AE & \textbf{53.78} & \underline{63.39} & 63.70 & \underline{54.71} & \textbf{55.55} & \underline{53.89} \\
& \textit{RobustDebias} & 56.05 & \textbf{58.67} & 61.74 & \textbf{53.79} & \underline{56.51} & \textbf{52.09} \\
\hline
\multirow{7}{*}{RoBERTa}
& ERM & 63.51 & 68.89 & 65.10 & 62.44 & 67.93 & 65.04 \\
& Group & 54.89 & 63.23 & 59.65 & 54.78 & 60.48 & 58.10 \\
& Topic CVaR & 67.99 & 67.90 & 70.19 & 65.01 & 63.87 & 68.10 \\
& TopK & \underline{56.69} & 61.75 & 58.90 & 56.58 & 62.58 & 55.65 \\
& TopK-Group & 58.19 & \underline{57.63} & 58.56 & \underline{57.66} & 56.82 & 56.84 \\
& TopK-AE & 58.95 & 57.89 & \underline{54.47} & 57.67 & \underline{55.80} & \underline{53.75} \\
& \textit{RobustDebias} & \textbf{53.19} & \textbf{54.76} & \textbf{52.09} & \textbf{54.34} & \textbf{53.99} & \textbf{53.61} \\
\hline
\multirow{7}{*}{DistilBERT}
& ERM & 36.71 & 64.23 & 63.40 & 36.49 & 64.47 & 61.05 \\
& Group & 58.23 & 66.09 & \underline{56.95} & 57.88 & 61.82 & 56.59 \\
& Topic CVaR & 25.14 & 39.67 & 40.88 & 24.99 & 39.23 & 39.37 \\
& TopK & \underline{46.87} & \underline{45.12} & 57.79 & \underline{44.70} & 42.96 & 34.24 \\
& TopK-Group & 39.83 & 62.84 & 35.56 & 39.59 & 62.15 & \underline{56.79} \\
& TopK-AE & 57.61 & 44.50 & 60.42 & 57.26 & \underline{43.79} & 58.18 \\
& \textit{RobustDebias} & \textbf{47.01} & \textbf{46.87} & \textbf{56.24} & \textbf{47.67} & \textbf{47.80} & \textbf{55.01} \\
\hline
\multirow{7}{*}{ALBERT}
& ERM & 68.89 & 34.21 & 31.09 & 65.20 & 34.27 & 31.80 \\
& Group & 44.09 & 57.88 & 39.51 & 39.05 & 56.14 & 40.88 \\
& Topic CVaR & 63.44 & 67.09 & 74.56 & 60.97 & 64.89 & 65.07 \\
& TopK & 39.79 & 59.84 & 40.12 & 39.76 & 58.04 & 41.09 \\
& TopK-Group & 40.39 & 37.79 & \underline{59.16} & 38.99 & 36.66 & \underline{57.62} \\
& TopK-AE & \underline{44.87} & \underline{56.69} & 61.26 & \underline{42.29} & \underline{53.99} & 59.40 \\
& \textit{RobustDebias} & \textbf{55.09} & \textbf{47.65} & \textbf{45.79} & \textbf{52.26} & \textbf{46.35} & \textbf{47.47} \\
\hline
\end{tabular}}
\caption{CrowS-Pairs: Metric scores for each demographic across all models with phase finetuned Split models. Closer to 50 is better (\textbf{Best}; \underline{Next Best})}
\label{table:crow-all-models-6bph}
\end{table}









\end{document}